\documentclass[11pt,letterpaper]{article}
\usepackage[top=1.0in, bottom=1.0in, left=1.0in,
right=1.5in]{geometry}
\newif\ifpdf\ifx\pdfoutput\undefined\pdffalse\else\pdfoutput=1\pdftrue\fi
\ifpdf \usepackage[pdftex]{graphicx} \pdfcompresslevel=9 \else 
\usepackage{graphicx} \fi
\usepackage{amsfonts,eucal,amsbsy,amsopn}
\usepackage{url}
\usepackage{times}
\usepackage[sort&compress]{natbib}
\usepackage{natbibspacing}
\bibpunct{(}{)}{;}{a}{,}{,}
\usepackage[usenames,dvipsnames]{color}
\usepackage{fancyhdr}

\fancypagestyle{firststyle}{
  \fancyhead{}
  \fancyfoot{}
  \fancyfoot[CO,CE]{{\thepage} of \pageref{lastpage}}
  \fancyhead[CO,CE]{\emph{\textcolor{Gray}{Draft 3 (July 15, 2012); comments welcome.}}}
}
\usepackage[font=small,labelfont=bf]{caption}

\DeclareMathOperator*{\argmax}{arg\,max}
\newcommand{\marginalia}[1]{\marginpar{\raggedright\footnotesize
  \textcolor{black}{$\triangleleft$ #1}}}
\newcommand{\ignore}[1]{}

\title{Adversarial Evaluation for Models of Natural Language}
\author{Noah A. Smith\thanks{The ideas in
  this paper were first presented in a talk at the 
NAACL-HLT Workshop on Inducing Linguistic Structure, June 7, 2012,
invited by organizers Phil Blunsom, Trevor Cohn, and Joao Gra\c{c}a;
many thanks to members of that audience for helpful comments and
thoughtful questions.
Thanks also to those with whom conversations about these
matters have sharpened my thinking:
Fred Jelinek,
Dan Bikel, Jordan Boyd-Graber, Alex Clark, Jaime Carbonell, Chris Dyer, Kevin Gimpel, Geoff Gordon, Lori Levin, Adam
Lopez, Andr\'{e} Martins, Behrang Mohit, Brendan O'Connor, Kemal Oflazer,
Philip Resnik, Bryan Routledge, Nathan Schneider, Yanchuan Sim, Amber
Wilcox-O'Hearn, and Dani
Yogatama.  Errors are, of course, the responsibility of the author.} \\
School of Computer Science \\ Carnegie
    Mellon University \\ Pittsburgh, PA 15213, USA \\
    \texttt{nasmith@cs.cmu.edu}}
\date{}

\begin{document}
\maketitle
\thispagestyle{firststyle}

\begin{abstract}
  We now have a rich and growing set of modeling tools and algorithms
  for inducing linguistic structure from text that is less than fully
  annotated. In this paper, we discuss some of the weaknesses of our
  current methodology. We present a new abstract framework for
  evaluating  natural language processing (NLP) models in general and unsupervised NLP models in
  particular. The central idea is to make explicit certain adversarial
  roles among researchers, so that the different roles in an
  evaluation are more clearly defined and performers of all roles are
  offered ways to make measurable contributions to the larger goal.
  Adopting this approach may help to characterize model successes and
  failures by encouraging earlier consideration of error analysis. The framework can be instantiated in a variety of ways, simulating
  some familiar intrinsic and extrinsic evaluations as well as some
  new evaluations.
\end{abstract}

\section{Introduction}

This paper presents a new approach to evaluating computational models
of natural language, based on adversarial roles performed by different
researchers or their models.
We begin in \S\ref{se:current} with a brief review of current
evaluation strategies in NLP.  We then turn to coupled adversarial evaluations inspired
by perplexity (\S\ref{se:perplexity}) and the traditional roles of
linguists (\S\ref{se:zellig}).  The two-performer setup is formalized
in \S\ref{se:formal}.  We then consider the origins of the data and
growing awareness of the importance of \emph{context} on language use
(\S\ref{se:context}) and provide a three-performer setup, in which a
third performer manages data selection (\S\ref{se:formal2}).  We
close with a few open questions (\S\ref{se:open-questions}).

\section{Current Evaluation Strategies in NLP}
\label{se:current}

At present, NLP models are primarily evaluated in three ways:
intrinsic evaluations, in which model predictions 
are compared to manually produced ``gold-standard'' output; extrinsic
evaluations, in which output
is passed downstream to an application whose performance can in turn
be evaluated; 
algorithm competitions with predefined, formal criteria for success
(less typical in NLP but used in some related areas);
and (for probabilistic models) perplexity evaluations,
in which the model is used to
assign a likelihood score to unseen data, and this score is compared
to other models' scores.  We assume the reader is familiar with these styles
of evaluation and consider their strengths and weaknesses in turn.  

\subsection{Intrinsic Evaluations: \textsc{MatchLinguist}}  \label{se:match-linguist}

Models can be evaluated by comparing their predictions on input data to which they have never previously been exposed to the predictions of human experts (known as ``gold-standard'' linguistic annotations).
 This is the dominant intrinsic evaluation
approach in NLP.
In \citet{smith-05c}, we introduced the term ``\textsc{MatchLinguist}'' to
refer to the task of automatically reproducing gold-standard
linguistic annotations. 

The strength of intrinsic
evaluations is that, once gold-standard annotations are provided, they
can be reused forever.  Once a scoring algorithm is agreed upon, many
researchers can evaluate their models on the same data, making
quantitative comparison easy.  Unfortunately, this makes it difficult
to draw conclusions about how well performance results generalize to
other linguistic samples (e.g., in different genres, topics, dialects,
languages, etc.).  Indeed, some have conjectured that long-term reuse
of an annotated test dataset can lead to community-wide
``overfitting'' to the pecularities of the data and the 
conventions used in annotating it.  (\citealp{wagstaff12}, recently expressed concern over this trend in the field of machine learning, emphasizing the gap between such datasets and ``real world'' problems.) The recent trend of developing new, small testing
datasets, often in a range of languages or genres, helps to alleviate
this problem (e.g., the CoNLL depedency parsing shared tasks; \citealp{buchholz-06}).

A major problem with intrinsic evaluations is that they assume the
phenomenon of interest is already well-enough understood that
linguistic experts have identified the best representation and trained annotators to produce it accurately.  Anyone who has
worked on an annotation project, however, knows that interaction with
the data always leads to evolution within the annotation scheme.
Intrinsic evaluation further commits the fallacy that human evaluations are
worthy of replication.  Annotators are only human,
and we have very restricted ways of evaluating them (e.g.,
inter-annotator agreement).  These annotator evaluations are often
incomplete, ignoring major factors like the kind and amount of
training annotators have been subjected to and the learning
curve of the annotators.  A model that succeeds at the
\textsc{MatchLinguist} task can be said to have reproduced
what a particular set of annotators, with a particular kind of
training, on a particular kind of data, within a particular amount of
time, would generate on the test set.  Our view is that drawing
stronger conclusions about the quality of such a model may be too bold.

Even if we accept human annotations as correct, the intrinsic
evaluation strategy only permits comparison between models that
produce similar output.  We cannot use it to test two divergent
theories of linguistic structure without introducing more automation
or manual effort.\footnote{For an example of this problem, consider the
  literature on unsupervised part-of-speech tagging.  A whole range of
  evaluation scores for this problem exist, each proposing a different
  way of dealing with the
  incommensurability of categories that come from linguistic
  annotators or various unsupervised learning models.}
  We also cannot use it to test
models of the interaction of different levels of structure (e.g.,
morphology and syntax), unless all levels of interest are part of the
human annotation effort or all models make use of the same
preprocessing mechanisms, whether manual or automatic.  The more we
are forced to incorporate pre- and post-processing in order to
evaluate our models, the more narrow our claims must be.

Finally, the cost of employing linguistic experts to annotate data is
often cited as one of the main motivating factors for unsupervised
NLP.  When we consider that any annotation project used to evaluate an
unsupervised NLP model has been necessarily limited in how much data
annotators could annotate and how many iterations they could make over
the data, it seems that using these cost-constrained annotations as
a gold-standard to match is misguided.  Unsupervised automatic
learners should be able to reason about far more data far more
consistently; should matching what resource-limited humans can do
really be our aim?  As noted by Alex Clark (personal communication), linguists really perform two tasks in creating an annotated dataset, and the labor is likely divided.  One task is defining the formalism:  what is the set of analyses that are possible for each input?  The other is selecting the correct one for each input.  Supervised NLP models focus only on the latter task, while unsupervised NLP models may perform both tasks, depending on the underlying assumptions.

Though it has not been consistently articulated
this way, and not all NLP researchers are likely to agree, perhaps we
should consider the goal of \emph{doing linguistics}---describing and
explaining the phenomena of human language, or defining formalisms, as above---in ways that unaided humans cannot. Indeed, as long as the mark of success for an unsupervised linguistic learner is to closely match what annotation scheme designers and annotators believe they already know about language, we cannot claim that \textsc{MatchLinguist}-evaluations of unsupervised NLP models have anything to do with advancing the scientific study of language.

\subsection{Extrinsic Evaluations:  Passing the Buck}

Extrinsic evaluations are attractive because they allow NLP modelers to make
claims of ``usefulness'' about their models.  Real-world applications
that use NLP models of various kinds include machine translation systems, search
engines, information extraction systems, and question answering
systems.  Insofar as evaluation of these systems' quality is
uncontroversial, there is little to be said against an argument for
the usefulness of a model whose output improves the downstream state
of the art.  Unfortunately, system evaluation remains fraught with
debate for most of these downstream applications.

There is also a practical concern:  evaluating an NLP model in a
downstream system requires the ability to incorporate that model's
functionality within such a system.  The
open-source versions of such applications do not generally provide a
``plug-and-play'' architecture for linguistically annotated input, and
if they do, there are strong assumptions about what kind of annotation
is to be provided.  How to use a particular linguistic annotation
within any given application is itself a research question.  Further,
as in intrinsic evaluations, models that make use of different kinds
of representations will not generally be comparable in downstream
applications, since much will depend on the process of incorporating the
annotations into the application.  Finally, applications change fast.
There is value in having stable mechanisms to compare models; yet the
community tends to show little interest in
performance gains in a downstream application that is no longer the
state of the art, since the results may not generalize to newer,
better systems.

All is not hopeless, and extrinsic evaluations should continue to
provide evidence for model quality.
However, we are not
optimistic that downstream applications can serve as the primary
evaluation mechanism for NLP models, due to these challenges of access
and stability.

\subsection{Algorithm Competitions}

Some research agendas lead naturally to the design of competitions in which a well-defined problem is stated formally and benchmarks to test an algorithm's success are constructed by experts.  A notable example is the Omphalos competition \citep{starkie04}, in which competitors constructed context-free grammar learning algorithms.  Theoretical and practical matters were taken quite seriously; datasets were designed to be provably sufficient for identifying the language but beyond the capabilities of the state of the art at the time.  Some of the benchmarks were constructed around natural language phenomena.  Though our proposed adversarial evaluations here seek to drive natural language modeling research, not formal language acquisition, much of what was done by the Omphalos designers and competitors focused on the construction of negative examples, which also play a key role here.\footnote{We gratefully acknowledge Alex Clark for bringing this to our attention.}

\subsection{Perplexity}

The idea of using model likelihood on test data to compare
probabilistic models arose in the speech recognition community, where
it was applied to the evaluation of language models.  It provides a
simple way to compare any models that properly assign
probability mass to linguistic data.  Perplexity evaluations were mostly
abandoned in the 1990s when it became clear that perplexity reductions
did not correlate with word error rate reductions on speech
recognition tasks.  In general, it is widely known that having a good
probablistic model of data need not have anything to do with
having a model that performs well in intrinsic or extrinsic
evaluations.  Perplexity's use as a scientific tool is less
controversial, though it is not widespread or widely accepted in
computational linguistics today, with the possible exception of the
Bayesian topic modeling subcommunity \citep[see,
e.g.,][]{blei-03}.\footnote{Even in that community, the usefulness of
  perplexity has been brought into question; \citet{chang09} found
  that topic models with better perplexity may infer relatively less
  semantically meaningful topics.}

There are also some key difficulties.  First, only \emph{probabilistic} models define
perplexity scores.  While probabilistic modeling has many attractions, requiring that researchers adopt that framework in order to compare with other work is unnecessarily exclusionary.   Second, two models' perplexity scores are only
comparable if they define exactly the same event space.  In practice,
this means prior agreement on the vocabulary and on handling of
out-of-vocabulary terms.  This must be done with great care, because
events that are assigned very low probability by a model can have a large effect
on the model's perplexity score.  (Assigning zero, in particular, leads to
infinite perplexity.  Perplexity offers no way to rank two models that
have infinite perplexity, no matter how sharp their differences on the
non-zero-probablity instances.)  Focusing on perplexity can lead to over-attention to smoothing algorithms, the details of which may be less important in large-data settings \citep{brants-07}.   And finally, many models in use today
involve latent structures, so that perplexity---which is calculated by
marginal inference---can only be calculated approximately. Conclusions
based on approximate perplexity comparisons, with each researcher
deciding on his or her own approximations, are suspect at best.

\section{Improving Perplexity:  Claude, the Chooser}
\label{se:perplexity}

Perplexity, in its most general form, requires the performer to
define a probability distribution $p$ over some event space $\mathcal{X}$.
During the evaluation period, a series of events $\langle x_1,
\ldots, x_N\rangle$ are assumed to be drawn i.i.d.~from the ``true''
distribution, which we
denote by $p^\ast$.  The estimated perplexity is given by:
\begin{equation}
\exp_2 {\displaystyle \left( - \frac{1}{N} \sum_{n=1}^N \log_2 p(x_n)
  \right)} \hphantom{\sum} \approx \exp_2 {\displaystyle \left( - \sum_{x\in \mathcal{X}} p^\ast(x) \log_2
    p(x) \right)}
\end{equation}
\noindent (On the right is the true perplexity, approximated on the
left using a test sample.  We use $\exp_2 (a)$ to more clearly denote $2^a$.)
As we have noted, the event space $\mathcal{X}$, is often an infinitely
large discrete space (e.g., the space of strings for a given
alphabet).  Performers must assign nonzero probability to all $x \in
\mathcal{X}$, and be able to compute that probability, to be able to compete.

Suppose we replace the calculation with a choice between two elements
of $\mathcal{X}$, $x$ and $y$, the former being true data from
$p^\ast$, and the latter being contrived or synthesized data.  These
two elements would be presented in random order, hiding the provenance
of each.
Performers might use probabilistic
models to make this decision (e.g., choosing $x$ iff $p(x) > p(y)$,
and choosing $y$ otherwise),
but they need not do so.  Any approach could be applied to
make the choice.  Insofar as the ability to distinguish real data from
contrived data is of scientific or practical interest, we would prefer
a model with greater average accuracy on this binary task.

We will conflate the engineer of such a model and the model itself,
calling both ``Claude.''\footnote{After Claude Shannon (1916--2001),
  the father of information theory.}  Claude takes as input two
instances from $\mathcal{X}$---sentences in the language modeling
case---denoted by $x$ and $y$.  $x$ is assumed to be drawn from a
true linguistic sample, and the other, $y$, to be contrived by an adversary who is given access to $x$, and
whom we call ``Zellig.''\footnote{After Zellig Harris (1909--1992),
  linguist and methodologist of science.}  We will return to Zellig in \S\ref{se:zellig}, for now taking
it for granted that Zellig's role can be meaningfully performed.

We remark on a few observations about this task:
\begin{itemize}
\item The accuracy score is easy to calculate, objective, and does not hinge on
  any human input beyond the choice of the test data $\langle x_1,
  \ldots, x_N\rangle$ and the machinations of Zellig to construct
  confusion instances $\langle y_1, \ldots, y_N\rangle$.
\item Comparing Claude to an alternative, competing model ``Chloe'' is straightforward,
  regardless of their internal operations and representations of the
  data.  In particular, they need not use the same theory (or any
  theory), and they need not use probabilistic models.  They only need
  to distinguish true data from contrived data.
\item Zellig's role is crucial.  If Zellig creates $y$ through some
  ``safe,'' trivial operation on $x$ (e.g., replacing common words
  with common words of the same syntactic category, or, worse, just
  copying $x$; or selecting instances from a corpus that closely
  approximates the same $p^\ast$ whence $x$ is drawn), then for reasonably large $N$ no Claude will be able
  to achieve better than 50\% accuracy.  On the other hand, if Zellig
  is built to be completely ignorant (e.g., sampling from a character $n$-gram
  distribution), it should be easy for any Claude to achieve very high performance.
\end{itemize}

\section{Zellig, Transformer of Data}
\label{se:zellig}

It quickly becomes clear that the quality of a Zellig must be defined in
terms of contemporary Claudes (and vice versa).  A good Zellig, in short, is one that
stumps the Claudes of the day, but not all to the same degree.  In
some sense, Zellig is like an extrinsic task evaluation, except that
rather than taking a model of language's predictions as output, we imagine that it
challenges that model to be aware of phenomena that hold in real
linguistic data but not in corrupted versions of those data.
We consider next three kinds of Zelligs, each suggesting a different research goal that is interesting regardless of its role in stumping Claudes.

\subsection{Human Zellig}

A human linguistic expert who has a theory of language might manually
corrupt a real linguistic utterance $x$ to create an ill-formed similar
utterance $y$.  The minimal pair then represents a prediction of Zellig's
theory.  Supporting evidence for the theory might come from a fluent (or native-speaker)
human Claude, who selects which of the two utterances is ill-formed.
In this setup, if Claude performs well (across many instances),
Zellig's theory is given some credence.  This is simply an experiment for testing claims
about well-formedness in natural languages.  A more useful variation
might compare Zellig to another expert, Zelda, to see whose theory
better predicts native speaker judgments.

This alters the role of a linguist from
annotator (\S\ref{se:match-linguist}) to creative illustrator.  Rather
than constructing theories that seek to account ``horizontally'' for \emph{all}
phenomena at a particular level of description in a natural language
(e.g., syntax), the linguistic expert is free to consider ``vertical'' interactions
among any levels at all that are appropriate to identifying selected
phenomena in a language.

\subsection{Model of Language Zellig}

It is, of course, a small step to imagine that human Zellig would write a
program to perform the $x \mapsto y$ transformation.  Indeed, 
many existing models of language can be used to construct a Zellig.
For example, a probabilistic language model might be queried to find a
string that has high probability and low (but nonzero) ``distance''
from $x$:
\begin{equation}
y = \argmax_{x' \in \mathcal{X} : 1 \le \Delta(x', x) \le \delta} p(x')
\end{equation}
\noindent where $\Delta(x', x)$ might be the Hamming distance or some
other metric on strings.\footnote{In past work evaluated within the \textsc{MatchLinguist} paradigm, we considered functions that generated large ``neighborhood'' sets of strings similar to $x$ but perturbed in ways expected to corrupt linguistic quality.  The corrupting function was heuristic and served the definition of a ``contrastive'' objective function for learning. See, for example, \citet{smith-05c}.}
  A model of linguistic constraints might identify the
constraints holding in $x$, then make a change that violates one of
them.  A little-discussed property that any computational model of
language might be expected to have is the ability to produce instances in violation
of the underlying theory.   \marginalia{Key idea:  using
  computational models of language to corrupt real instances.}
We propose that explicit construction of
algorithms for this use-case is motivated as a new way to validate models in
computational linguistics, and further may lead to new insights about
computational models of linguistic phenomena.
Note further that the \emph{same model}, if it provides algorithms for
both kinds of queries, might serve as Zellig or Claude in different evaluations.

\subsection{Text-Generating System Zellig}
\label{se:generation}

Another kind of Zellig can be constructed as an NLP system that
generates text as output.  For example, suppose that each $x_n$ is a
sentence in English that was translated (by a human) from French.
Assume $x_n$ comes packaged with metadata, which we will denote $m_n$,\marginalia{We'll return to metadata below.}
which is comprised of the original French sentence.  If Zellig is a
machine translation system, then $y_n$ will be an automatic
translation of French sentence $m_n$. Another scenario might
  consider question answering:  $m_n$ is a question, $x_n$ its human
  answer, and $y_n$ the answer from a system.  Two versions might be
considered here, one where Claude observes $m_n$ (encouraging
evaluation of adequacy of translation or correctness of question answering), and one where it is
kept hidden (judging only fluency).

\marginalia{We might have started the discussion by talking
about evaluation of Zellig, rather than Claude!}The setup therefore provides a new way to perform system evaluations,
exploiting models of language that seek to pass distinguishability
tests.  

\section{Defining the Two-Performer Evaluation}
\label{se:formal}

\begin{figure}
\centering \includegraphics{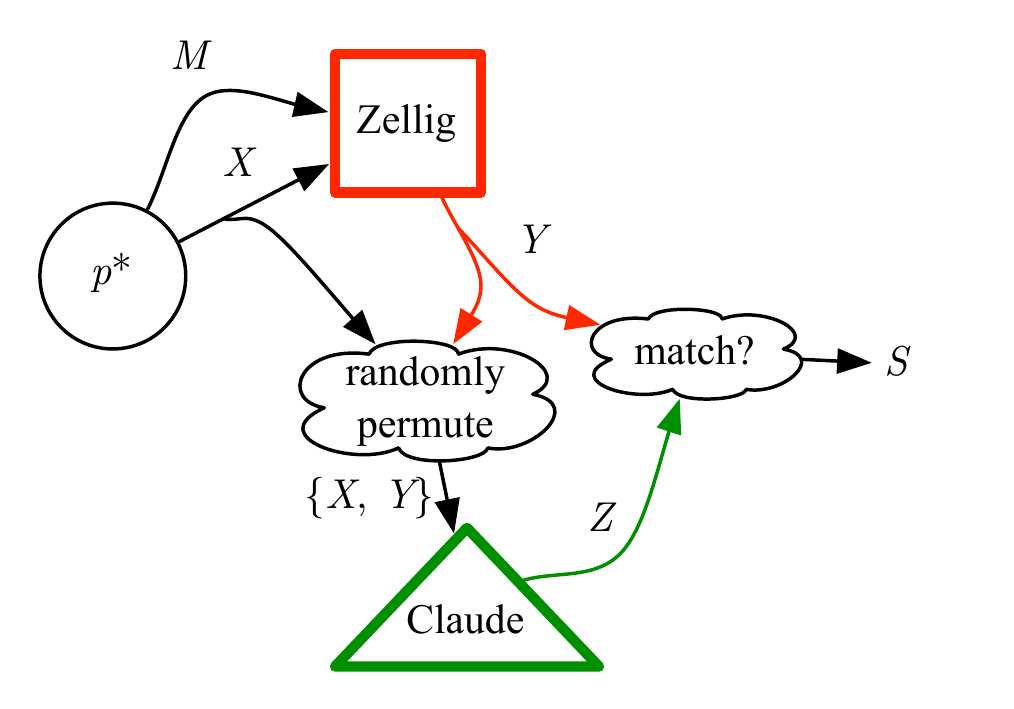}
\caption{The basic setup for the $Z$-task and the $C$-task (see
  \S\ref{se:formal}).  Colors and shapes were chosen to match the
  graphics used in the talk on June 7, 2012. Here $M$ is not made
  available to Claude, but in some settings it might be.  \label{fig:basic-setup}}
\end{figure}

A formal definition of the basic adversarial evaluation follows;
Figure~\ref{fig:basic-setup} provides an illustration.

We assume a random source of linguistic instances, the probability distribution $p^\ast$.  Each
instance  $x_n$ is drawn from $p^\ast$.  Optionally, $p^\ast$ defines
a joint distribution over the instance random variable $X$ and a
metadata random variable $M$.  We assume that instances are generated
at fixed periodic intervals of length $t$.

The evaluation involves two performers, Zellig and Claude, who---though in
an adversarial relationship---are not in direct competition with each
other.  Zellig can be compared to other performers of the $Z$-task,
defined below,
using the same sample and Claude. Claude can be compared to other
performers of the $C$-task, defined below, using the same sample and
Zellig.

On  \marginalia{$Z$-task defined.}  each iteration (indexed by $n$), Zellig takes $x_n$ (and optionally
$m_n$) and constructs an object $y_n$, purportedly from the support of
$p^\ast$.  If there is metadata $m_n$, then Zellig should seek $y_n$
that is well-paired with $m_n$.  It is to Zellig's advantage to choose
$y_n \not = x_n$.
Zellig must perform all necessary computation within time $t$, before
the next iteration.\footnote{If Zellig does not provide $y_n$ in time
  $t$, a reasonable default is to set $y_n = x_n$, thereby giving a
  free point to Claude.}  This is the $Z$-task.

On \marginalia{$C$-task defined.} each iteration, the pair $(x_n,
y_n)$ is permuted uniform-randomly and presented to
Claude. Claude must guess which element is the false instance $y_n$; all necessary
computation must be within time $t$, before the next
iteration.\footnote{If Claude does not make a choice in time $t$, then
  a uniform-random choice should be made.} We
denote by $z_n$ Claude's choice.
(It is helpful to think of Claude as one step behind Zellig, so that
at an arbitrary point in time, Zellig is generating $y_n$ and Claude
is guessing between $x_{n-1}$ and $y_{n-1}$.)  This is the $C$-task.

Over $N$ instances, the score is defined by:
\begin{equation}
S = \frac{1}{N} \sum_{n=1}^N \boldsymbol{1}\{ z_n = y_n \} 
 \hphantom{\sum} \approx \mathbb{E}_{p^\ast(X)}[ \boldsymbol{1}\{Z = Y\} ] 
\label{eq:score}
\end{equation}
For two competing performers of the $Z$-task, the one achieving the
\emph{lower} $S$ is the winner; this is a Zellig who more successfully
deceived Claude.  For two competing performers of the
$C$-task, the one achieving the \emph{higher} $S$ is the winner; this
is a Claude who has more successfully distinguished real data from
corrupted data.

\paragraph{A connection to cryptography.} \label{crypto-paragraph}  
Our $C$-task bears some similarity to a particular notion of security
of an encryption scheme called ``indistinguishability of encryptions'' \citep{bellare98}.%  Our $C$-task bears some similarity 
% to a particular notion of security of an encryption scheme involving a ``chosen ciphertext attack'' \citep{blum90}.
\footnote{We gratefully acknowledge Amber Wilcox-O'Hearn for introducing this connection in personal communication.}  To highlight the similarity, we will employ Claude as an analyst and Zellig as an encryption oracle within a particular kind of attack known as ``chosen ciphertext.''  In this evaluation, Claude chooses two plaintext messages and sends both to Zellig.  Zellig chooses one at his discretion, encrypts it, and sends the ciphertext to Claude.  Zellig succeeds---and the scheme declared secure---if  Claude can do no better than chance at guessing which message Zellig chose.  Returning to our linguistic setup, for Zellig to have this ability would be evidence for  ``strong linguistic knowledge.''   Much like the notion of ``provable security,'' ``strong linguistic knowledge'' should perhaps be regarded with skepticism; a scheme can be considered secure only until it is broken, and a Zellig remains respectable only as long as the state-of-the-art Claudes cannot consistently perform well on his output.    (Note, however, that the evaluation we have proposed does not allow Claude any control over the inputs to Zellig.)

\paragraph{Spam detection.}  Several members of the audience at the June 7 talk noticed the similarity between Zellig and Claude's activities and the adversarial relationship between spammers and spam detection software.  Of course, spammers are constrained by the speech act they seek to execute through the message $y$, and they have no analogue to ``$x$,'' unlike Zellig.  Successful spam detection systems presumably exploit this (and metadata $m$) heavily.  (For an interesting recent discussion of spammer strategy, specifically considering the linguistic choices involved, see \citealp{herley12}.)

\paragraph{Game theory.}  We have deliberately avoided discussing the proposed evaluation in game-theoretic terms.  On reading a draft of this proposal, economist Bryan Routledge exclaimed, ``I want this data,'' seeing long-term transcripts of the choices by Claudes and Zelligs as inherently interesting in studying the dynamics of ``co-evolution'' in evolutionary game theory \citep{weibull95}.  We leave this possible point of exploration for future work.

\section{On Context}
\label{se:context}

The reader may have noticed the introduction of a largely
underspecified element, metadata $m_n$ on each iteration, in
\S\ref{se:generation}.  The importance of \emph{context}---encoded in
our setup as metadata---to interpretation
and generation of language has been noted with increasing intensity in
recent discourse about NLP. Context can include well-studied variables
like the dialect or genre in which language is produced, or simply
``co-text'' (a term used in a recent presentation by Graeme Hirst to refer to nearby text), or farther-removed
representations of the situation in which the text arose.\footnote{Social
media platforms offer a rich set of possibilities for the last of
these, since messages are broadcast from an identifiable individual
with a history, to a set of identifiable individuals connected to her,
at a known timestamp, etc.}
In recent research efforts, we and others have made the prediction of
contextual information from text a task of its own, often predicting
future contextual variables from text in a forecasting setup (e.g., \citealp{kogan-09}).

A \marginalia{Key idea: rejection of $p^\ast$; data selection methods
  should be critically evaluated.} strong statement of the importance
of context is to claim that $p^\ast$ is such an over-simplification of
reality as to be useless.  Indeed, all corpora currently used to
construct and evaluate NLP systems come with some description of the
provenance of the text.  The community already views
the construction of contextualized linguistic resources as a valuable
research effort; what we lack are frameworks for objectively
evaluating the quality of such datasets or their relevance to
scientific or engineering efforts, and the early incorporation of this
information into our models.  The evaluation framework proposed here
offers a first step toward imposing the same kind of rigorous
evaluation on data selection methods as on data modeling methods.

We noted that one role of linguistic experts in this framework is as a
human Zellig performer who contrives
corrupted instances $y$ from observed linguistic instances $x$.  By introducing a third performer, called
``John,''\footnote{After John Sinclair (1933--2007), a corpus linguist.}
we propose another role for linguistic experts---the curation of
linguistic datasets with contextual descriptions, and the construction
of systems to perform this task.  We have reached a time when raw text
data is available in massive amounts, often with metadata as
\emph{objets trouv\'{e}s}; the collection and further description of
such data (adding to the metadata) naturally feeds the adversarial
evaluations we have proposed so far.

A useful by-product of John's performance is the generation of
metadata that enables error analysis.  Many researchers desire understanding of the kinds of
systematic ``mistakes'' that NLP systems make, but we have very few methodological tools for this kind of characterization.  Many researchers resort to fine-grained statistics on errors or selection of illustrative
examples, but these do little to show the way forward for future
research. Correlating errors to well-defined phenomena marked in
metadata may be a more useful tool in gaining an understanding of
what a given model (performing either as Zellig or as Claude) solves
or does not solve.

\section{Defining the Three-Performer Evaluation}
\label{se:formal2}

\begin{figure}
\centering \includegraphics{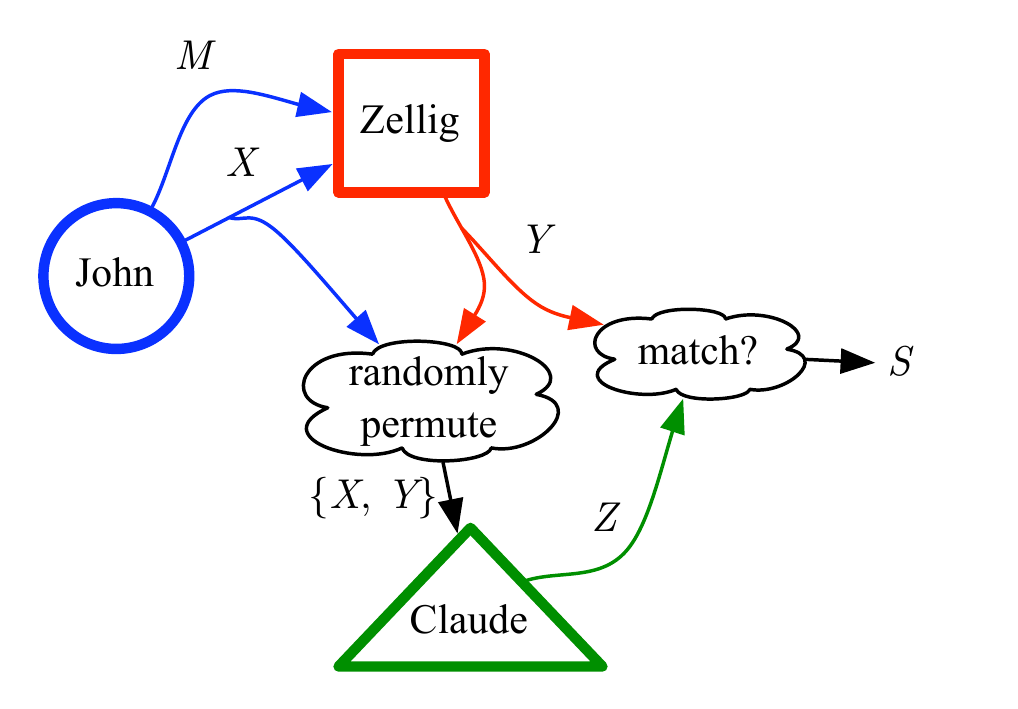}
\caption{The full setup for the $J$-task, $Z$-task, the $C$-task (see
  \S\ref{se:formal2}). Colors and shapes were chosen to match the
  graphics used in the talk on June 7, 2012.  Here $M$ is not made
  available to Claude, but in some settings it might be. \label{fig:full-setup}}
\end{figure}

The \marginalia{$J$-task defined.} full three-performer adversarial evaluation  is illustrated in
Figure~\ref{fig:full-setup}.  Claude and
Zellig are exactly as before (\S\ref{se:formal}).  We introduce
performer John, who replaces ``$p^\ast$'' as a source of pairs $(x_n,
m_n)$.  John's contribution is to collect 
 data that reveal phenomena on contextualized natural language, or to
 implement algorithms that collect such data.
Researchers performing this task---the $J$-task---are expected to
justify the relevance of the selection method for scientific
exploration and/or engineering NLP systems.  Further, John can be
compared to another $J$-task performer ``Jennifer'' by selecting a Zellig-Claude pair and measuring
the scores $S$ as achieved before.  John is said to outperform
Jennifer, given Zellig and Claude, if Zellig's task has been made
harder with John's data than with Jennifer's (i.e., a higher score results).

\begin{table}
\centering \begin{tabular}{l|l|l}
\multicolumn{1}{c}{Name} &
\multicolumn{1}{c}{Task} &
\multicolumn{1}{c}{Evaluation} \\
\hline
$C$-task (Claude) & distinguish data from non-data & high $S$\\
$Z$-task (Zellig) & generate corruptions of data & low $S$\\ 
$J$-task (John) & select data to exemplify phenomena of interest &
high $S$ \\
\hline
\end{tabular}
\caption{Summary of the three performers.}
\end{table}

Any evaluation on the three tasks ($C$-task, $Z$-task, and $J$-task) requires
the presence of the other two performers.
Just as evaluations on multiple datasets with different properties are
often used to make stronger arguments in NLP, an evaluation can be
strengthened by considering a range of other performers.  For example,
in evaluating Claude, we might compare his performance against a
baseline Chloe on a range of evaluations $\mathcal{Z} \times
\mathcal{J}$, where $\mathcal{Z}$ is a set of existing Zelligs and
$\mathcal{J}$ a set of existing Johns.  Higher-level analysis can be
performed by relating $S$ to properties of the Zelligs or Johns.  The wider the range of other performers, the more confident we can be that an evaluation result is not due to idiosyncrasies.

The discussion has been fairly abstract; we have deliberately avoided
making assumptions about what kinds of resources a performer might
have access to constructing the algorithm or model to perform a task.
The original idea was conceived out of skepticism toward evaluations
for \emph{unsupervised} NLP models, but we believe the framework is appropriate
regardless of the level of supervision.  Indeed, the framework
forces us to differentiate two different kinds of supervision:
\begin{enumerate}
\item Supervision from the \textsc{MatchLinguist} perspective, in
  which expert annotations are provided in support of the task.
\item Supervision within the task:
\begin{itemize}
 \item In the $C$-task, observations of tuples $(m_n, x_n, y_n)$, with
   $X$ and $Y$ labeled as
   such (rather than randomly permuted), for a given John-Zellig pair.
\item In the $Z$-task, observations of a given Claude's choices in response to
  the generated $y_n$, given each $(m_n, x_n)$ pair from a given John.
\item In the $J$-task, observations of the generated $y_n$ from a
  given Zellig and $z_n$ from a given Claude in response to each
  $(m_n, x_n)$ produced by the performer.
\end{itemize}
We call a round of evaluation $n$ ``transparent'' from the perspective of
a given performer if that performer can see the other performers'
actions clearly in the round.\footnote{An extremely adversarial
  variant of the evaluation might allow all performers some
  transparent rounds.  While entertaining, we believe such a scenario
  begins to lose attraction as a way to objectively compare systems in a highly controlled, understandable setting.}
\end{enumerate}

From each performer's perspective, \textsc{MatchLinguist} supervision,
though perhaps useful, is indirect.  From a learning perspective, it
is the second kind of supervision that is expected to give the most
information.  
Given the framework, it is easy to explain
 supervised, semi-supervised, and unsupervised versions of
 the evaluation.  We must simply specify the schedule of
 observations---transparent and non-transparent---that occur before
 evaluation takes place.  A few interesting cases include:
\begin{itemize}
\item zero observation rounds prior to evaluation;
\item a fixed number of transparent rounds (``supervised'') prior to evaluation;
\item a fixed number of transparent rounds followed by a fixed number
  of non-transparent rounds (``semi-supervised'') prior to evaluation;  or
\item a fixed number of non-transparent rounds (``unsupervised'')
  prior to evaluation.
\item Orthogonal to all of the above, performers might adapt their
  performance during the evaluation's non-transparent rounds.
\end{itemize}
Regardless of which framework is selected, the explanation of any
performer should clarify what resources were used to construct it, and
how, as in current NLP research.  For frameworks that involve adaptation, reporting how the score changes over time (e.g., as a time series) would be useful for comparing convergence rates.

Finally, we suggest again that any of the roles might be played by humans.  Such an exercise might be useful in establishing human ``upper bounds'' (risking the problems underlying \textsc{MatchLinguist}), or in training any of the performers.  An example suggested by Amber Wilcox-O'Hearn is a human Claude who provides supervision for a supervised learner Zellig.

\section{Open Questions}
\label{se:open-questions}

\paragraph{Collusion.}  We have assumed that, for clarity's sake, collusion
among performers should not be allowed.  However, collusion between
any pair might lead to more challenging evaluations for the third
performer and might be worth considering.  %See, for example, the connection to chosen ciphertext attacks from cryptography on page~\pageref{crypto-paragraph}.

\paragraph{Cheating.}  Is it possible to cheat?  Validity of
evaluations in this framework may rest on limiting the resources
available to some performers.  We believe, for
example, that it is possible to cheat if John does not have access to
external data sources unavailable to Zellig and Claude.

\paragraph{Should we do it?} A reasonable concern, raised by Dan Bikel, is that the collective actions of a community of Johns, Zelligs, and Claudes might not lead to improved models of natural language.  The ideas laid out here are intended to refocus our evaluations so as to build better models of the phenomena inherent in language.  Yet it is not hard to imagine that researchers would collectively over-attend to $S$ (Equation~\ref{eq:score}) and lose sight of those phenomena.  

We therefore propose a modest start.  A few straightforward John, Zellig, and Claude performers should be publicly released, perhaps through an API allowing inspection of all data, algorithms, and scores.  If the API is extended to allow new performers to join in and test performance, we conjecture that the adversarial framework will begin to be used to provide evidence for the quality of newly developed models.  Whether this evidence is judged meaningful by the research community will depend, of course, on the particulars.  We believe, though, that a critical assessment of our evaluation practices, and the introduction of some new ones, can only benefit future research.

\ignore{
\section{Possible To-do List}
\begin{itemize}
\item RTE challenge -- early data selection -- tagging instances with
phenomena (not a representative sample) ... McCartney's thesis, table
7.4 [who told me about this?]
\item  ICML position paper ... ``ML not connected enough to applications'' --
discussion on Fernando Pereira's G+, testing web-scale IE
\item Zweig and Burges workshop paper; construction of the MSR paraphrase corpus; pseudo word sense disambiguation -- all cases where a system was used to construct a task
\item talk about building a platform with an API for anyone to
play any role, either synchronously with contemporaneous performers or
asynchronously with data from the past and server-side performers?
\end{itemize}
}

\bibliographystyle{plainnat}
\bibliography{lsp}
\label{lastpage}
\end{document}